\documentclass[11pt]{article}
\usepackage{fancyhdr}
\usepackage{amssymb}
\usepackage{amsmath}
\usepackage{mathrsfs}
\usepackage{color}
\usepackage[toc,page]{appendix}
\usepackage{url}
\usepackage{marginnote}
 
\usepackage{graphicx}
\graphicspath{{Figs/}}
\DeclareGraphicsExtensions{.eps,.pdf,.jpg,.gif}
\def\figdir{./figs}

\usepackage[linesnumbered, ruled]{algorithm2e}

\def\here{
 \[\vcenter{
\medskip
\hbox to\hsize{\hfill\footnotesize here $\downarrow$}
\smallskip
\hrule
\medskip}\]}

\font\fiverm=cmr5
\def\R{{\bf R}}

\def\feat{C\mskip -10mu\lower-2pt\hbox{\fiverm 1}\,}
\def\Feat#1{C\mskip -10mu\lower-2pt\hbox{\fiverm #1}\,}
 
\def\dash---{\thinspace---\hskip.16667em\relax} 

\def\I{I}
\def\O{O}
\def\H{H}
\def\pa#1{\mathop{\rm pa}\nolimits(#1)}
\def\ch#1{\mathop{\rm ch}\nolimits(#1)}
\def\(#1){[\hbox{$\mkern1mu\thickmuskip=\thinmuskip#1\mkern1mu$}]} 
\def\loss{v}
\def\|#1|{\hbox{ \includegraphics{\figdir/BP-#1.mps}}}
\usepackage{epsfig}

\title{\bf  Backpropagation and \\ Biological Plausibility}
\author{Alessandro Betti, Marco Gori, and Giuseppe Marra\\
SAILab, University of Siena \\
}

\begin{document}

\thispagestyle{empty}

\maketitle

\begin{abstract}
	By and large, Backpropagation (BP) is regarded as one of the most important 
	neural computation algorithms at the basis of the progress in
	machine learning, including the recent advances in deep learning. 
	However, its computational structure has been the source of 
 	many debates on its arguable biological plausibility.
	In this paper, it is shown that when framing supervised learning in the Lagrangian
	framework, while one can see a natural emergence of Backpropagation,
	biologically plausible local algorithms can also be devised 
	that are based on the search for saddle points in the learning adjoint space
	composed of weights, neural outputs, and Lagrangian multipliers.
	This might open the doors to a truly novel class of learning algorithms where, because
	of the introduction of the notion of support neurons, the
	optimization scheme also plays a fundamental role in the construction of the
	architecture. 
\end{abstract}

\section{Introduction}
Backpropagation is considered to be biologically implausible. 
Amongsts others, Stefan Grossberg, stressed this fundamental
limitation by discussing the transport of the weights that is assumed in the algorithm.
He claimed that ``Such a physical transport of weights has no plausible physical interpretation.''~\cite{Grossberg87b}. Since then, a number of papers have contributed 
with many insights on the overcoming of this limitation, 
including recent contributions~\cite{DBLP:journals/corr/BengioLBL15},\cite{DBLP:journals/corr/ScellierB16}.

The ideas behind this paper nicely intercept Yan Le Cun's paper on a theoretical framework for Backpropagation that is based on a Lagrangian formulation of learning with the neural equations imposed as constraints~\cite{leCun_cmss89}. 
It is worth mentioning that he established the connection
by imposing the stationary condition of the Lagrangian, which is in fact the only case in which
one can restore the classic factorization of Backpropagation with the forward and backward terms.
\footnote{This connection 
was brought to our attention during a discussion with Yoshua Bengio on biological 
plausibility of Backpropation and related algorithms. We find that, because of the close
connections with the ideas proposed in this paper, this is a nice instance of ``Chi cerca trova, chi ricerca ritrova.''  (Ennio De Giorgi), that is quite popular amongst mathematicians. 
}

Instead of focussing on the derivation of Backpropation in
the  Lagrangian framework, in this paper we introduce a novel approach to learning 
that explores the search in the learning adjoint space that is characterized by the
triple $(w,x,\lambda)$ (weights, neuron outputs, and Lagrangian multipliers.)
According to the prescriptions on the discovery of the minimum, we search
for saddle points in this space. It turns out that the gradient descent in the 
variables $(w,x)$ and the gradient ascent in the multipliers $\lambda$ give rise to 
algorithms that are fully local. This confers them a biological plausibility 
that is not enjoyed in Backpropation.  

An important consequence of the proposed scheme is that it
 is very well-suited for a network gradual building. 
Algorithms can be constructed   to build neurons that gradually satisfy the constraints imposed
by the training set while respecting their own underlying model. 
In a sense, at any stage of learning, the algorithm is characterized by the property
of  finding a perfect building (PB) of the neurons. 


\section{Lagrangian formulation of supervised learning}
\label{LagrForm-section}
Our model is based on any directed graph $D=(V,A)$, where $V$ is the set of
vertices of $D$ and $A$ is the multiset of arcs of the graph.
We denote with $\I$ the set of input
neurons, with $\O$ the set of the output neurons and with $\H = V \setminus
(\I \cup \O)$
the set of hidden neurons.
Suppose furthermore $\sigma(a):= a\, \(a>0)=(a)_{+}$ \footnote{
  We are using here the Iverson's notation: Given a statement
  $A$, we set $\(A)$ to $1$ if $A$ is true and to $0$ if $A$ is false.}.
Given a training set $\mathscr{L}=\{\,(x_{\kappa i}, y_{\kappa i})
\mid i\in O, \quad \kappa=1,\dots,\ell\}$,
learning is formulated as the problem of determining $w^\star$ as a solution
of the problem

\begin{equation}
\label{LearningFormulation}\vcenter{
\halign{#&\quad$#$\hfil\cr
minimize &\sum_{\kappa=1}^\ell \big( \sum_{i \in \O}
	\loss(x_{\kappa i},y_{\kappa i}) 
	+  \sum_{j \in \H}
	\alpha_{\kappa j} |x_{\kappa j}| \big)\cr
        subject to & x_{\kappa i} - \sigma \big(\sum_{j \in \pa{i}} w_{ij} x_{\kappa j} \big)= 0, \quad  i\in H\cup\O,\quad\kappa =1,\ldots,\ell,\cr}}
\end{equation}
with $\alpha_{\kappa j} \geq 0$.
Clearly, once we determine $w^{\star}$, the corresponding $x^{\star}$
is determinded from the neural constraint.
The error function $\loss(x_{\kappa i},y_{\kappa i}) +  \sum_{j \in \H}
\alpha_{\kappa j} |x_{\kappa j}|$, which is accumulated all over the patters,  
is minimized under  the neural architectural constraints.
Here, $\loss\colon \R\times\R\to\R^+$
is a loss function, while the weighted $\ell_1$ regularization term favors 
sparse solutions that correspond with neural pruning.  
The corresponding Lagrangian is $L(w,x,\lambda)
=\sum_{\kappa=1}^\ell L_\kappa(w,x,\lambda)$ with
\begin{align}
\begin{split}
	L_\kappa(w,x,\lambda) &= \sum_{m}\Bigl(
	\loss(x_{\kappa m},y_{\kappa m})\, \(m\in \O) 
	+ \alpha_{\kappa m} |x_{\kappa m}|\, \(m\in \H)\\
	& +\lambda_{\kappa m} \bigg(x_{\kappa m} - 
	\sigma \Big(\sum_{r \in \pa m} w_{mr} x_{\kappa r} \Big) \bigg)
        \,\(m\in \H\cup \O)\Bigr)
\end{split}
\end{align} 
Now, let $\sigma^{\prime}_{\kappa i} = \sigma^{\prime} \bigl(\sum_{r \in\pa i} w_{r i} 
x_{\kappa r}\bigr)$
be. In order to analyze the consequence of the saddle point condition
of the Lagrangian, we calculate
\begin{align}
\label{BPfactorization}
  \partial_{w_{ij}} L &=  \sum_{\kappa=1}^\ell\partial_{w_{ij}} L_\kappa
	 = - \sum_{\kappa \in \mathscr{L}} \lambda_{\kappa i}  \sigma_{\kappa i}^{\prime} x_{\kappa j} \\
\label{RecursiveBP}	 
	  \partial_{x_{\kappa i}} L &=   \partial_{x_{\kappa i}} L_\kappa=
	  [i \in \O]  \big(
	   \partial_{x_{\kappa i}} \loss + \lambda_{\kappa i}
	   \big) \\  \nonumber
	   &\phantom{=\partial_{x_{\kappa i}} L_\kappa=} + [i \in \H] \bigg(\alpha_{\kappa i} \frac{x_{\kappa i}}{|x_{\kappa i}|} + \lambda_{\kappa i}
	   - \sum_{r \in\ch i} \sigma^{\prime}_{\kappa r} w_{r i} \lambda_{\kappa r} \bigg)\\
  \partial_{\lambda_{\kappa i}} L &= \partial_{\lambda_{\kappa i}} L_\kappa=
	 x_{\kappa i} -  \sigma \Big( \sum_{j \in \pa i} w_{i j} x_{\kappa j}\Big).
\end{align}
Once we set $\alpha_{\kappa j} = 0$, if we impose 
the stationarity conditions on the Lagrangian
 $\partial_{x_{\kappa i}} L=0$ and $\partial_{\lambda_{\kappa i}} L=0$
then we recognize in Eq.~(\ref{BPfactorization}) and Eq.~(\ref{RecursiveBP})
the classic structure of the Backpropagation algorithm. In particular,
Eq.~(\ref{BPfactorization}) expresses the gradient with respect to the
weights by the factorization of the forward and backward terms, while in the 
Eq.~(\ref{RecursiveBP}), when imposing  $\partial_{x_{\kappa i}} L=0$,  
yields the backward update of the delta term, which is
interpreted (with negated sign) by the corresponding Lagrangian multiplier.
In particular, we have 
\begin{equation}
	\lambda_{\kappa i} = \left\{
	\begin{array}{ll}
	 - \partial_{x_{\kappa i}} \loss & i \in \O \\ 
	\sum_{r \in \ch i} \sigma^{\prime}_{\kappa r} w_{r i} \lambda_{\kappa r}
	& i \in \H.
	\end{array}
\right.
\end{equation}

Finally, $\partial_{\lambda_{\kappa i}} L=0$ clearly reproduces the 
neural constraints. This theoretical framework of Backpropagation was 
early proposed in a seminal paper in~\cite{leCun_cmss89}, at the down of the connectionist wave.


\section{Plausibility in the learning adjoint space}
The idea behind the proposed approach is inspired by the basic differential multiplier method (BDMM)
\cite{DBLP:conf/nips/PlattB87}.
Instead of using the Lagrangian approach as a framework for Backpropagation,
one can think of searching for
saddles points by the following learning algorithm, which updates
the parameters according to

\medskip
\begin{algorithm}[H]	
	Initialize $\lambda, x, w$;\\
	\Repeat 
	{stopping criterion}
        {
        		\For{$\kappa=1\  \emph{\KwTo}  \ \ell$} {
	  		\For{$i=1\  \emph{\KwTo}  \ n$} {
        				$x_{\kappa i} \leftarrow x_{\kappa i} - \mu_{x}\partial_{x_{\kappa i}} L$; \label{XUpd}\\
        				$\lambda_{\kappa i} \leftarrow \lambda_{\kappa i} 
				+ \mu_{\lambda} \partial_{\lambda_{\kappa i}} L$; \label{LambdaUpd}\\
			}
        	    	}
		\For{$i=1\  \emph{\KwTo}  \ n$} {
	   		\For{$j=1\  \emph{\KwTo}  \ |\pa i|$} {
        	    			$w_{ij} \leftarrow w_{ij} - \mu_{w}   \partial_{w_{ij}} L$;  \label{WeightUpd} 
        	  		}
		}
        }
         \caption{Saddle point discovery}
\end{algorithm}

\medskip
This is a batch-model learning algorithm which operates in the {\em
learning adjoint space} defined by the triple $(w,x,\lambda)$.  The
parameters can be randomly initialized, though in the following we
will discuss the initialization $\lambda \leftarrow 0$.  Notice that
we need to minimize with respect to the variables $(w,x)$, so as we
perform gradient descent according to lines~(\ref{WeightUpd})
and~(\ref{XUpd}).  On the opposite, the Lagrangian multipliers are
updated by gradient ascent as stated in line~(\ref{LambdaUpd}). It is
worth mentioning that the parameter updating of $x_{\kappa i}$ and
$\lambda_{\kappa i}$ exhibits the same batch-mode structure adopted
for $w_{ij}$, but they are kept separate to emphasize the structure of
the variables $(w,x,\lambda)$ in the adjoint learning space.

Now, we give some more details on  the algorithm with the purpose
of proving that, unlike Backpropation, it is fully local. 

\medskip
\begin{algorithm}[H]	
	Initialize $\lambda, x, w$; $\partial_x L\gets 0$, $\partial_\lambda L\gets 0$; $f \leftarrow 0$\\
	\Repeat 
	{stopping criterion}
        {
        		\For{$\kappa=1\  \emph{\KwTo}  \ \ell$} {
	  		\For{$i=1\  \emph{\KwTo}  \ n$} {
				\eIf{$i \in \O$} 
        				{$\partial_{x_{\kappa i}} L \leftarrow \partial_{x_{\kappa i}} \loss + \lambda_{\kappa i}$}
        				{\For{$r=1\  \emph{\KwTo}  \ \ch i$} {
         			$\partial_{x_{\kappa i}} L \leftarrow \partial_{x_{\kappa i}} L
        				- \sigma^{\prime}_{\kappa r} w_{r i} \lambda_{\kappa r}$
                                      }
                                      {$\partial_{x_{\kappa i}} L\gets \partial_{x_{\kappa i}} L+\lambda_{\kappa i}+\alpha_{\kappa i}
                                        x_{\kappa i}/|x_{\kappa i}|$}
				}
				\For{$j=1\  \emph{\KwTo}  \ \pa i$} {
					$f_{\kappa i} \leftarrow f_{\kappa i} + w_{ij} x_{\kappa j}$
				}
				$\partial_{\lambda_{\kappa i}} L \leftarrow x_{\kappa i} - \sigma(f_{\kappa i})$; \label{LBUpd}\\
        				$x_{\kappa i} \leftarrow x_{\kappa i} - \mu_{x}\partial_{x_{\kappa i}} L$; \label{RXUpd}\\
        				$\lambda_{\kappa i} \leftarrow \lambda_{\kappa i} 
				+ \mu_{\lambda} \partial_{\lambda_{\kappa i}} L$; \label{RLambdaUpd}\\
	   			\For{$j=1\  \emph{\KwTo}  \ \pa i$} {
        	    			$w_{ij} \leftarrow w_{ij} - \mu_{w}   
				 \lambda_{\kappa i}  \sigma_{\kappa i}^{\prime} x_{\kappa j}$;  \label{RWeightUpd} 
				}
        	  		}
        		}
        }
         \caption{Local computation of saddle points}
\end{algorithm}

\medskip
We begin noticing that the weights are updated in line~(\ref{RWeightUpd}),
according to Eq.~(\ref{BPfactorization}). As already noticed, this has the classic factorization 
structure of Backpropagation. Interestingly, the output $x_{\kappa i}$ is not determined by 
the forward computation as in Backpropagation algorithm, since $x_{\kappa i}$ evolves
according to the updating line~(\ref{RXUpd}). As we can see, the updating ends as 
$\partial_{x_{\kappa i}} L \rightarrow 0$, which corresponds with the satisfaction of 
Backpropagation backward step (see Eq.~(\ref{RecursiveBP})). As already noticed,
in this case, the Lagrangian multipliers $\lambda_{\kappa i}$ can be interpreted as the
Backpropagation delta error. The computation of the Lagrangian multiplier $\lambda_{\kappa i}$,
that is used in the factorization  Eq.~(\ref{BPfactorization}), follows line~(\ref{RLambdaUpd})
and exhibits a dual structure with respect to $x_{\kappa j}$. It requires to accumulate 
$\partial_{\lambda_{\kappa i}} L$ that is determined in line~(\ref{LBUpd}). The multipliers
ends up its updating as $\partial_{\lambda_{\kappa i}} L \rightarrow 0$, which corresponds with
satisfaction of the neural constraints defined by Eq.~(\ref{LearningFormulation}).
When looking at the overall structure of the algorithm, one promptly realizes that the outer loops
on $\kappa$ and on $i$ drive the computation over all examples and neurons.

\parshape 7 0pt 23pc 0pt 22pc 0pt 21pc 0pt 21pc 0pt 21pc 0pt 21pc 0pt \hsize
Interestingly, for any pair $(\kappa, i)$ the factors $x_{\kappa j}$ and $\lambda_{\kappa i}$
are computed by involving $\ch i$ and $\pa i$, respectively (see side figure). This corresponds exactly with
the operations needed in Backpropation for the backward and forward step, but the remarkable
difference is that these computations take place at the same time in a local way on each neuron
$i$ without needing the propagation to outputs. The algorithm is $\Theta(|{\cal A}| \cdot \ell)$,
where ${\cal A}$ is the set of arcs of the graph, that is
it exhibits the same optimal asymptotical property of Backpropagation.
\smash{\raise 2pc\rlap{\kern-7pc \vbox{\|1|\smallskip \hbox{\box0}}}}

This analysis reveals the full local structure of the algorithm for the discovery of saddle points,
a properly that, unlike Backpropagation, makes it biologically plausible.

Notice that the extension to on-line and mini-batch modes requires a sort of prediction of
$x_{\kappa i}$ and $\lambda_{\kappa i}$, since when we move from one iteration to the other, 
unlike for batch mode, no variable is allocated to keep the corresponding value. 
This is an interesting issue, which seems to indicate that the biological plausibility that is
gained by the proposed scheme is either restricted to the possibility of storing all the variables
according to the batch scheme or to the presence of an inherent temporal structure 
that is found in most perceptual tasks.




\section{Support neurons and support examples}
Let us begin with the assumption that there is no regularization ($\alpha_{\kappa i}=0$)
and that the learning process has ended up into the condition $\loss=0$.
From Eq.~(\ref{RecursiveBP}) 
we can promptly see that $\forall k$, $\forall i:$ $\lambda_{\kappa i}=0$. 
In other words, the neural constraints do not 
provide any reaction. Whenever this condition holds true, we say 
that we are in front of {\em straw neurons}.
In case any of the (hidden) neurons is removed in such a way to keep
the end learning condition $\loss=0$ true, we fully appreciate the property that
we are in front of a needless neuron. Of course, if we continue pruning 
the network, after awhile the end learning condition $\loss=0$ will be violated.
Interestingly, as this happens, all neurons likely turn into {\em support neurons},
that are characterized by $\lambda_{\kappa i} \neq 0$. 
This trick transition is somehow indicating the ill-position of learning when
formulated as in Eq~(\ref{LearningFormulation}). On the other hand, also
in SVM, the support vectors emerge because of the presence of regularization. 

If $\alpha_{\kappa i} \neq 0$ then we suddenly see the emergence of a new
mathematical structure that reveals the essence of a network building plan 
that is driven by the parsimony principle. As we can see from Eq.~(\ref{RecursiveBP}),
the end of learning condition $\loss=0$ does not correspond with the nullification
of the Lagrangian multipliers anymore. While for $i \in \O$ we have
$\forall k$, $\lambda_{\kappa i} = 0$, for $i \in \H$ we have
\begin{equation}
	\alpha_{\kappa i} \frac{x_{\kappa i}}{|x_{\kappa i}|} + \lambda_{\kappa i}=0.
\end{equation}
The partition based on the nominal satisfaction of the 
condition $x_{\kappa i} = 0$ at the end of learning allows us to distinguish
between straw from support neurons, since whenever $x_{\kappa i} = 0$
we have $\lambda_{\kappa i}=0$. Hence, a learning process that drives to
$\loss=0$ with some straw neurons suggests that we can remove them 
and rely on support neurons only.


\section{Conclusions}
This paper gives an insight on the longstanding debate on the unlikely biological plausibility of Backpropation. It has been shown that the formulation in the learning adjoint space under the Lagrangian formalism leads to fully local algorithms that naturally emerge when searching
for saddle points. The approach can be extended to on-line and mini-batch mode learning by keeping the weights as the state of learning process. The adoption of $\ell_{1}$ regularization makes also possible to appreciate the role of support and straw neurons, that can be pruned so as the learning process also contributes to the construction of the architecture.  Finally, the approach can also naturally extended to any recurrent network for sequences and graphs by expressing the corresponding constraints induced by the given data structure.


\begin{center}
{\bf \Large Acknowledgements}
\end{center}
We thank Yoshua Bengio for discussions on biological plausibility,
which also bring me back to closely related studies on a theoretical framework for Backpropagation by Yan Le Cun~\cite{leCun_cmss89}.


%
%
%

\bibliography{corr,nn}
\bibliographystyle{plain}
\end{document}